\documentclass[sigconf]{acmart}
\usepackage{colortbl}
\usepackage{color}
\usepackage{latexsym}
\usepackage{amsmath}

\usepackage{courier}
\usepackage{graphicx}
\usepackage{multirow}
\usepackage{booktabs}
\usepackage{amsfonts}
\usepackage{verbatim}
\usepackage{pgfplots}
\usepackage{wrapfig}
\usepackage{balance}

\usepackage{enumerate}
\usepackage{amssymb,mathrsfs}
\usepackage{amssymb}
\usepackage{array}
\usepackage{pgfplotstable}
\usepackage{pgf}
\usepackage{bm}
\usepackage{colortbl}
\usepackage{soul}
\usepackage{xspace}
\usepackage{graphicx}
\usepackage{enumitem}
\usepackage{hyperref}
\usepackage{graphicx}
\usepackage{subfigure}
\usepackage{adjustbox}
\usepackage{arydshln}
\usepackage{multirow}

\AtBeginDocument{
  \providecommand\BibTeX{{
    \normalfont B\kern-0.5em{\scshape i\kern-0.25em b}\kern-0.8em\texttteX}}}

\setcopyright{acmcopyright}
\copyrightyear{2023}
\acmYear{2023}
\acmDOI{XXXXXXX.XXXXXXX}

\acmConference[CIKM '23]{the 32nd ACM International Conference on Information and Knowledge Management}{October 21-25, 2023}{University of Birmingham and Eastside Rooms, UK}

\acmPrice{15.00}
\acmISBN{978-1-4503-XXXX-X/18/06}

\begin{document}
\copyrightyear{2023} 
\acmYear{2023} 
\setcopyright{acmlicensed}\acmConference[CIKM '23]{Proceedings of the 32nd
ACM International Conference on Information and Knowledge
Management}{October 21--25, 2023}{Birmingham, United Kingdom}
\acmBooktitle{Proceedings of the 32nd ACM International Conference on
Information and Knowledge Management (CIKM '23), October 21--25, 2023,
Birmingham, United Kingdom}
\acmPrice{15.00}
\acmDOI{10.1145/3583780.3615093}
\acmISBN{979-8-4007-0124-5/23/10}

\title{Towards Spoken Language Understanding via Multi-level
Multi-grained Contrastive Learning}
\renewcommand{\shorttitle}{MMCL}

\author{Xuxin Cheng}
\email{chengxx@stu.pku.edu.cn}
\affiliation{
  \institution{School of ECE\\
  Peking University}
  \country{}
}

\author{Wanshi Xu}
\email{xwanshi@stu.pku.edu.cn}
\affiliation{
  \institution{School of ECE\\
  Peking University}
  \country{}
}

\author{Zhihong Zhu}
\email{zhihongzhu@stu.pku.edu.cn}
\affiliation{
  \institution{School of ECE\\
  Peking University}
  \country{}
}
\author{Hongxiang Li}
\email{lihongxiang@stu.pku.edu.cn}
\affiliation{
  \institution{School of ECE\\
  Peking University}
  \country{}
}
\author{Yuexian Zou}
\authornote{Corresponding author.}
\email{zouyx@pku.edu.cn}
\affiliation{
  \institution{School of ECE\\
  Peking University}
  \country{}
}

\renewcommand{\shortauthors}{Xuxin Cheng, Wanshi Xu, Zhihong Zhu, Hongxiang Li, \& Yuexian Zou}

\begin{abstract}
Spoken language understanding~(SLU) is a core task in task-oriented dialogue systems, which aims at understanding the user's current goal through constructing semantic frames. SLU usually consists of two subtasks, including intent detection and slot filling. Although there are some SLU frameworks joint modeling the two subtasks and achieving high performance, most of them still overlook the inherent relationships between intents and slots and fail to achieve mutual guidance between the two subtasks. To solve the problem, we propose a multi-level multi-grained SLU framework MMCL to apply contrastive learning at three levels, including utterance level, slot level, and word level to enable intent and slot to mutually guide each other. For the utterance level, our framework implements coarse granularity contrastive learning and fine granularity contrastive learning simultaneously. Besides, we also apply the self-distillation method to improve the robustness of the model. Experimental results and further analysis demonstrate that our proposed model achieves new state-of-the-art results on two public multi-intent SLU datasets, obtaining a 2.6 overall accuracy improvement on the MixATIS dataset compared to previous best models.
\end{abstract}

\maketitle

\section{Introduction}
\begin{figure}[tb]
\begin{minipage}[a]{1.0\linewidth}
  \centering
\centerline{\includegraphics[width=7.75cm]{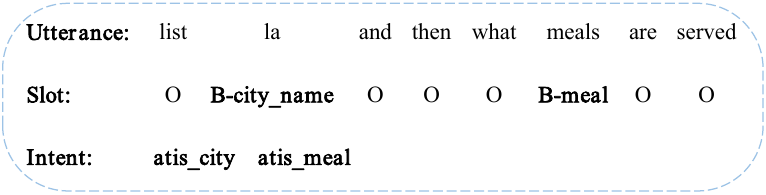}}
  \smallskip
\end{minipage}
\caption{An example of intent detection and slot filling.}
\label{fig:ex}
\end{figure}
In task-oriented spoken dialog systems, spoken language understanding (SLU) plays a crucial role in accurately comprehending the user's intent by constructing semantic frames \cite{young2013pomdp,qin2022gl,cheng2023accelerating,qin2023openslu}. The primary objective of SLU is to understand and extract relevant information from user's utterances, enabling the system to identify the user's current goal and provide appropriate responses. The SLU component leverages various techniques, including natural language processing (NLP) and machine learning algorithms to accomplish its subtasks. These techniques help the dialog systems to understand and extract relevant information from user's utterances accurately. By employing SLU in task-oriented spoken dialog systems, the system can effectively comprehend the queries, gather the necessary information, and respond to user's requests.

As shown in Figure \ref{fig:ex}, SLU usually includes two subtasks: intent detection and slot filling~\cite{tur2011spoken,zheng2022hit,dong2022pssat,zhu2024aligner2,zhuang2024towards}, where intent detection is a classification problem and slot filling is a sequence labeling problem. In real-world scenarios, a sentence often contains more than just a single intent. To this end, multi-intent SLU~\cite{xu2013convolutional,kim2017two,cheng2024towards} is explored, and \cite{gangadharaiah2019joint} makes the first attempt to jointly model multi-intent detection and slot filling in a multitasking framework.
Gradually, it becomes mainstream to study multi-intent SLU tasks, and more and more joint training models for multi-intent detection and slot filling are proposed.
Recently, \cite{qin2020agif} proposes an AGIF model for fine-grained multi-intent prediction via graph attention networks~(GAT)~\cite{velivckovic2017graph}, which adaptively integrates intent information into the autoregressive decoding process of slot filling. And \cite{qin2021gl} proposes GL-GIN, which builds a local slot-aware graph and a global intent-slot graph for each utterance, obtaining speedup and better accuracy compared to the autoregressive model. \cite{xing2022co} further proposes Co-guiding Net, which implements a two-stage framework achieving the mutual guidance between intent and slot. \cite{xing2022group} improves joint multiple intent detection and slot filling by exploiting label typologies and relations. 
\cite{Enhancing} proposes a novel approach to generate additional slot type-specific features to improve the accuracy. \cite{zhu2023acl} leverages label embeddings to jointly guide the decoding process of two tasks.

\begin{figure}[tb]
\begin{minipage}[a]{1.0\linewidth}
  \centering
\centerline{\includegraphics[width=7.75cm]{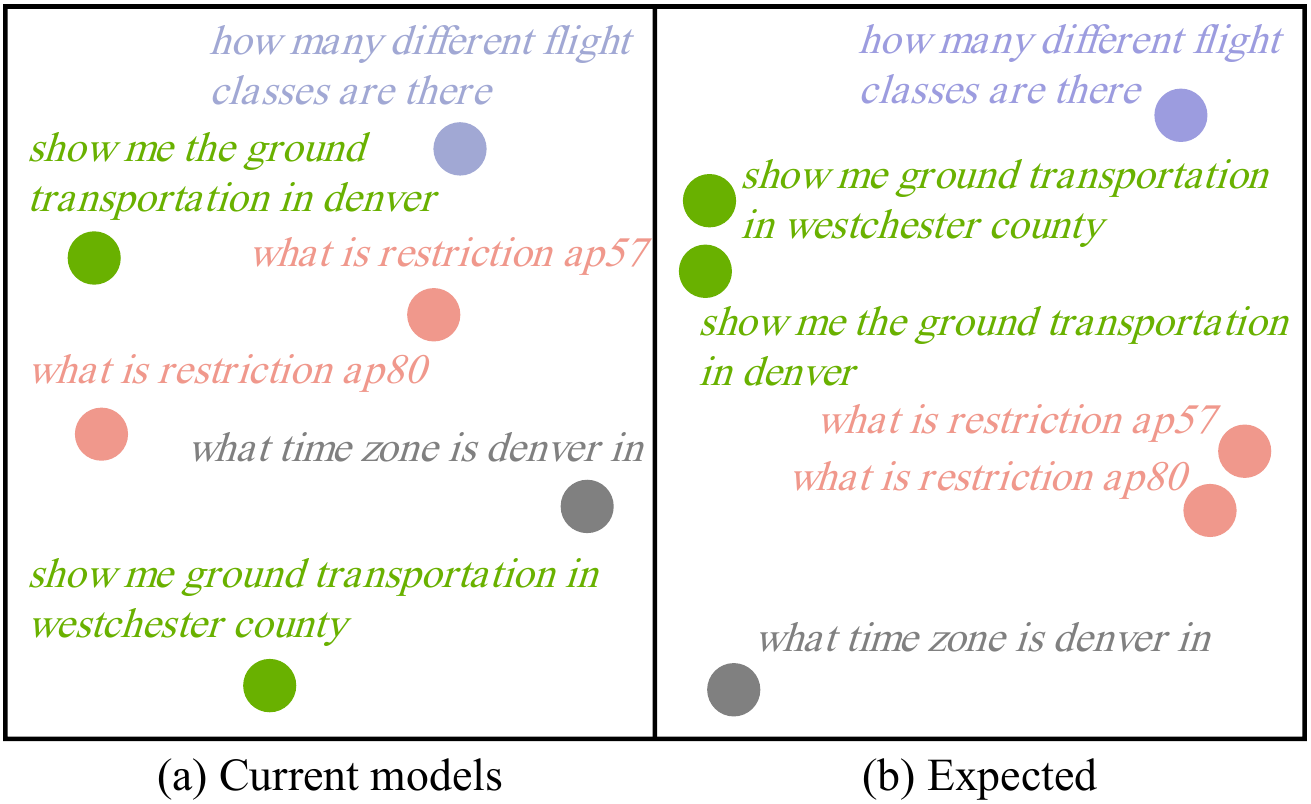}}
  \smallskip
\end{minipage}
\caption{Illustration of representations in utterance level for text~(projected to 2D), circles in the same color means they have similar semantics. (a): representations learned by existing models. (b): ideal representations that we expect, where semantically similar pairs should stay close to each other and semantically unrelated pairs should stay away from each other.}
\label{fig:example}
\end{figure}

Intent classification and slot filling are closely related and dependent on each other. The primary concerns of current and future proposed models are to enhance and leverage their relationship for joint training. Although many existing joint training models have achieved promising results, they still encounter two challenges:

(1)~\textbf{Fewer attempts are made to explore the inherent relationships in multi-intent SLU.} As shown in Figure \ref{fig:example}, the representations learned by existing models are distributed irregularly. Based on our observations, utterances with similar semantics and framing are more likely to have the same intents and slots. Therefore, it is beneficial to pull together features of semantically similar pairs and push away features of unrelated pairs. However, none of the previous works pay attention to this issue.

(2)~\textbf{Lack of guidance from slot to intent.} Although joint models for intent detection and slot filling have been widely used to improve utterance-level semantics through mutual enhancement between the two tasks, most of them only implement the guidance from intent to slot~\cite{guo2014joint, hakkani2016multi,chen2016syntax}. However, different slots can provide several relevant options for multiple intent detection. Therefore, the lack of guidance from slot to intent limits multiple intent detection and the joint task. Co-guiding Net~\cite{xing2022co} is proposed to address this problem, which implements a two-stage framework. Initial predictions of intent and slot are generated in the first stage, and they are used in the second stage for mutual guidance between intent and slot. However, despite the utilization of a margin penalty for each task in the second stage, it inevitably introduces error propagation.

To solve the aforementioned problems, we propose a novel model termed MMCL in this paper. MMCL implements a single-stage framework that applies margin-based contrastive learning to improve the performance of the model. Contrastive learning is applied in cross-lingual SLU tasks with substantial improvements. \cite{liang2022multi,qin2022gl} apply contrastive learning in cross-lingual SLU with the code-switching approach to exploit the semantic structure of the SLU task and facilitate explicit alignment. However, few works utilize contrastive learning for mutual guidance between slot and intent in monolingual SLU. Intuitively, utterances with similar semantics and framing usually have similar intentions and slots. Tokens with similar semantics can provide information for each other to generate multiple intentions and usually have the same slot. Moreover, intent detection is an utterance-level task while slot filling is a token-level task, so simply utilizing single-level contrastive learning is sub-optimal. Therefore, we implement contrastive learning in three levels, including utterance level, slot level, and word level. For the same reason, we both utilize fine and coarse granularity contrastive learning at the utterance level.

 For the first problem, contrastive learning is applied to learn the inherent hierarchical entailment relationships and mine the relationships between each unit at each level. For the second problem, slot-level and word-level contrastive learning can facilitate attention to similar slots in intent detection, which can provide guidance from slot to intent. By contrastive learning, the representations of words or slots with the same slot suffix will be closer. Since we joint model multi-intent detection and slot filling, these two tasks share utterance representation. As a result, our method could predict intents more accurately, which achieves the guidance from slot to intent. For the guidance from intent to slot, we feed the output of intent detection into the global graph interaction layer.
 
 Unlike Co-guiding Net~\cite{xing2022co}, MMCL is a single-stage framework that can effectively avoid the problem of error propagation. Moreover, in contrastive learning, we adopt margin-based similarity proposed in \cite{artetxe2019margin} which is reported to outperform simple cosine similarity. We also use multi-grained contrastive learning to further learn the structure of the utterance. Specifically, for utterance level, we use both coarse-grained contrastive learning and fine-grained contrastive learning. To improve the robustness of the model and prevent over-confidence, we add a self-distillation loss which minimizes KL divergence between the current prediction and the previous one. Note that Transformer~\cite{vaswani2017attention} is widely used in various tasks. However, most existing SLU models do not utilize Transformer. Therefore, for a fair comparison, we also do not use Transformer. Experiment results demonstrate that our MMCL  outperforms previous models and achieves state-of-the-art results on two benchmark datasets MixATIS and MixSNIPS~\cite{hemphill1990atis,coucke2018snips,qin2020agif}. Further analysis also
 verifies the advantages of our model. 
 
 The contributions of this work are three-fold:
\begin{itemize}[itemsep=2pt,topsep=0pt,parsep=0pt]
\item We propose a multi-level multi-grained contrastive learning framework for joint multiple intent detection and slot filling.
\item We apply a self-distillation method to improve the robustness of the model by minimizing the KL deviation between the current prediction and the previous prediction.
\item Experiment results on two benchmark datasets demonstrate that the MMCL achieves state-of-the-art performance.
\end{itemize}
\section{Related Work}
\subsection{Intent Detection and Slot Filling}
Due to the high correlation between the two tasks of intent detection~\cite{xia2018zero,chen2021dialogsum,cai2023dialogvcs} and slot filling~\cite{dong2023demonsf,dong2023multi,mao2023diffslu}, an increasing number of joint models~\cite{zhang2016joint,hakkani2016multi,goo2018slot,wu2020slotrefine,qin2021co,qin2021gl,xing2022co,xing2022group} are achieving excellent performance. 
To further investigate the relationship between Intent Detection (ID) and Slot Filling (SF), \cite{ liu2019cm, JointCapsule2019} introduce bi-directional networks. These works primarily focus on the hidden states of both tasks but lack the result information. To solve this limitation, \cite{qin2019stack} proposes a Stack-Propagation framework that utilizes the result information to refine SF with ID labels. Recognizing the impact of SF results on the ID task, \cite{clzICME2021} draws inspiration from \cite{yang2019} and presents a bi-feedback network called RPFSLU. RPFSLU guides the second round of predictions utilizing the results of the first round through representation learning. Although RPFSLU achieves satisfactory performance, it suffers from extended inference latency due to multiple rounds of SLU result prediction.
To accelerate the inference process, \cite{wu2020slotrefine} proposes a non-autoregressive approach called SlotRefine based on Transformer. This method successfully speeds up inference but encounters the problem of uncoordinated slots. To tackle this issue, \cite{cheng2021effective} proposes LR-Transformer, which incorporates a Layer Refined Mechanism and a specially designed auxiliary task. Building upon this work, \cite{clzTOIS} extends the LR-Transformer to multi-turn SLU tasks by introducing a Salient History Attention module. These advancements in the research contribute to a better understanding of the relationship between ID and SF. They propose novel techniques such as gate mechanisms, bi-directional networks, bi-feedback networks, non-autoregressive approaches, and refined models to improve the efficiency of SLU.

Recently, as the multi-intent problem gradually gains attention, some models based on graph attention mechanisms were gradually proposed. \cite{xu2013convolutional} and \cite{kim2017two} begin to study the Multi-Intent SLU. \cite{gangadharaiah2019joint} jointly learning SF and multiple ID via a multi-task framework. \cite{qin2020agif} proposes AGIF, which is based on a graph attention model to explicitly construct correlations between slots and intents. \cite{qin2021gl} designs a non-autoregressive global-local graph interaction network conducting parallel decoding for slot filling. Co-guiding Net~\cite{xing2022co} makes targeted improvements by designing a two-stage framework for mutual guidance between the two tasks, but introduces the error propagation problem, which is harmful to the model. Compared to previous works, the advantage of our model is that we circumvent the aforementioned problems by paying more attention to the mutual guidance without introducing error propagation.
\subsection{Contrastive Learning} 
\begin{figure*}[tb]
  \centering
  \includegraphics[width=\linewidth]{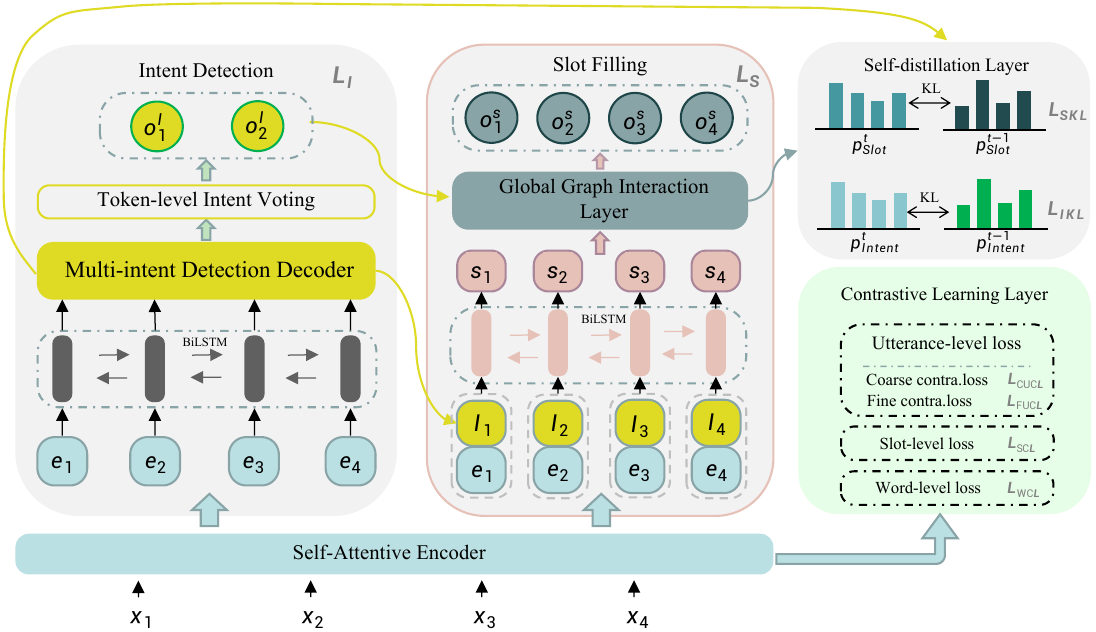}
  \caption{The main architecture of MMCL. We introduce margin-based multi-level multi-grained contrastive learning to explore the inherent relationships and achieve mutual guidance between intent and slot in SLU. And self-distillation method is applied to improve the robustness of the model and prevent over-confidence.}
  \label{fig:network}
\end{figure*}
Contrastive learning has gained significant attention in the field of unsupervised representation learning. Various methodologies and advancements have been proposed to improve the effectiveness and efficiency of contrastive learning. This section provides an overview of some key contributions in the related work of contrastive learning. \cite{bromley1993signature} introduces Siamese networks, which utilize shared weights to learn representations for pairs of input samples and are among the pioneering approaches in contrastive learning. The objective is to minimize the distance between similar pairs and maximize the distance between dissimilar pairs. Recently, with the development of pre-trained language models~\cite{li2023one,liu2024towards,gao2024self,zhang2024choose,liao2024baton,wang2024adashield,yao2024dual,yao2024multi,li2024ecomgpt,luo2024deem}, contrastive learning has been increasingly utilized.

Contrastive learning minimizes the distance between positive examples and maximizes the distance between negative examples~\cite{saunshi2019theoretical}. 
\cite{wolf2020transformers,devlin2019bert} first use contrastive learning to learn sentence representations in NLP. 
\cite{subramanian2018learning} proposes a Siamese LSTM network for learning sentence embeddings through contrastive learning. The model learns to maximize the similarity between positive sentence pairs and minimize the similarity between negative pairs.
\cite{chen2020improved} employs contrastive learning to enhance machine translation tasks, aiming to improve translation quality and learn more robust representations.
\cite{liang2022multi} proposes multi-level contrastive learning in cross-linguistic models to learn alignment in different language representation spaces. \cite{chang22c_interspeech} considers contrastive learning loss as a pre-training target to learn robust representations for errors in ASR. While most of the above models apply contrastive learning in cross-lingual or cross-modal tasks, our model applies contrastive learning in a monolingual environment by systematically formulating rules for generating positive and negative pairs at each level. In addition, we use margin-based similarity to replace cosine similarity as \cite{artetxe2019margin}.
\section{Approach}
\textbf{Problem Definition}
Given an input utterance $\boldsymbol{x}=\left(x_{1}, x_{2}, \dots, x_{n}\right)$, where $n$ is the length of $\boldsymbol{x}$. Multiple intent detection can be formulated as a multi-label classification task which outputs a sequence intent label $\boldsymbol{o}^I=\left(o_{1}^{I}, o_{2}^{I}, \dots, o_{m}^{I}\right)$, where $m$ is the number of intents in $\boldsymbol{x}$. And slot filling is a sequence labeling task that maps each utterance $x_i$ into a slot output sequence $\boldsymbol{o}^S=\left(o_{1}^{S}, o_{2}^{S}, \dots, o_{n}^{S}\right)$.

As illustrated in Figure \ref{fig:network}, we show our framework, including a self-attentive encoder~($\S\ref{Self-Attentive Encoder}$), a contrastive learning layer~($\S\ref{Contrastive Learning Layer}$), a token-level multi-intent detection decoder~($\S\ref{Multi-intent Detection Decoder}$), a global graph interaction layer~($\S\ref{Global Graph Interaction Layer}$), and a self-distillation layer~($\S\ref{Self-distillation Layer}$).

\subsection{Self-Attentive Encoder}\label{Self-Attentive Encoder}
Following \cite{qin2020agif,qin2021gl}, we utilize a shared self-attentive encoder with BiLSTM and a self-attention mechanism to produce initial hidden states containing basic semantics. BiLSTM reads the input sequence $\{{{x}}_{1}, {{x}}_{2}, \ldots, {{x}}_{n}\}$ forwardly and backwardly to capture the temporal dependencies, which can be formulated as follows:
\begin{equation}
\boldsymbol{h}_{i}=\operatorname{BiLSTM}\left(\boldsymbol{e}_{i}, \boldsymbol{h}_{i-1}, \boldsymbol{h}_{i+1}\right)
\end{equation}
where $\boldsymbol{e}_{i}$ is the word vector of $x_i$. The corresponding context-sensitive hidden states are denoted as $\boldsymbol{H} = \{\boldsymbol{h}_{1}, \boldsymbol{h}_2, \ldots, \boldsymbol{h}_{n}\}$. Self-attention mechanism is used to capture the context-aware features:
\begin{equation}
\boldsymbol{C}=\operatorname{softmax}\left(\frac{\boldsymbol{Q} \boldsymbol{K}^{\top}}{\sqrt{d_{k}}}\right) \boldsymbol{V}
\end{equation}
where $\boldsymbol{C}$ is the global contextual hidden states output by self-attention; $\boldsymbol{Q}$, $\boldsymbol{K}$ and $\boldsymbol{V}$ are matrices obtained by applying different linear projections on the input utterance word vector matrix.

Then we concatenate the output of BiLSTM and the self-attention mechanism as the output of the shared self-attentive encoder: 
\begin{equation}
\boldsymbol{E}=\boldsymbol{H} \mathop{||} \boldsymbol{C}
\end{equation}
where $\boldsymbol{E}$ = $\{\boldsymbol{e}_{1},\ldots,\boldsymbol{e}_{n}\}$ and $\mathop{||}$ denotes concatenation operation.

\subsection{Contrastive Learning Layer}\label{Contrastive Learning Layer}
\begin{figure*}[htb]
  \centering
  \includegraphics[width=\linewidth]{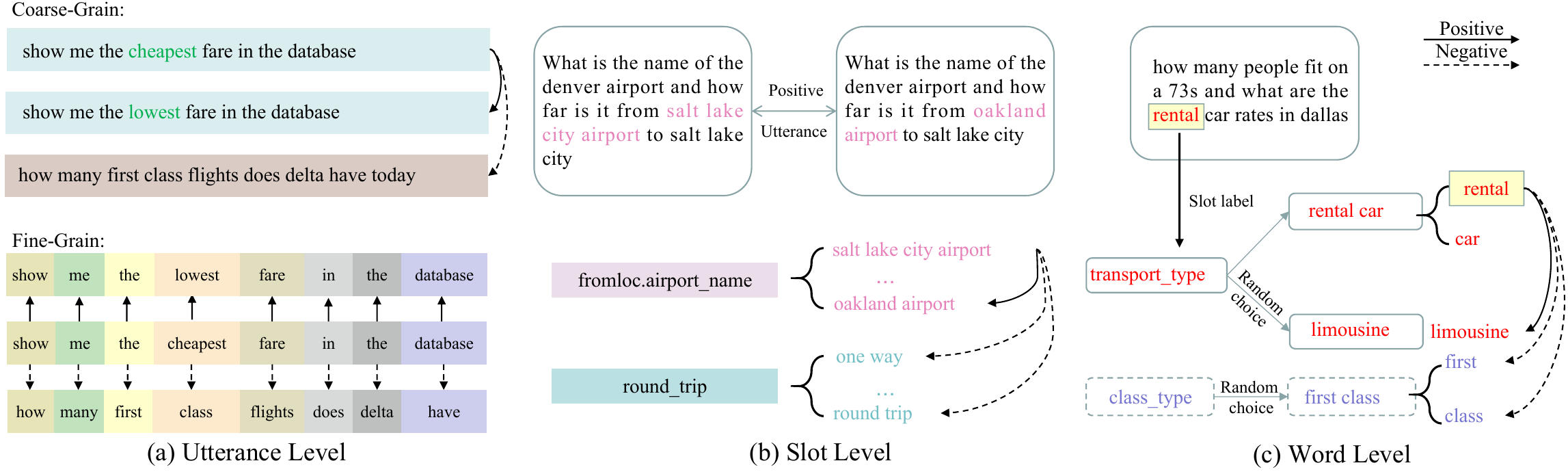}
  \caption{The illustration of margin-based multi-level multi-grained contrastive learning examples. For utterance level, the model implements coarse granularity and fine granularity contrastive learning simultaneously. }
  \label{fig:contrastive}
\vspace{2mm}
\end{figure*}

As shown in Figure \ref{fig:contrastive}, we utilize contrastive learning at three levels and two granularities with margin-based similarity. 

(1) \emph{Margin-Based Contrastive Learning}

Conventional contrastive learning suffers from the fact that the scale of cosine similarity is not globally consistent, which may lead to semantically similar pairs being pushed away incorrectly. To tackle this issue, we apply the margin-based similarity\cite{artetxe2019margin}:
\begin{equation}
\begin{split}
&\operatorname{sim}(x, y)=\cos(x, y)- \\ &\left(\sum_{z \in \mathrm{NN}_{k}(x)} \frac{\cos (x, z)}{2 k}
+\sum_{z \in \mathrm{NN}_{k}(y)} \frac{\cos (y, z)}{2 k}\right)
\end{split}
\end{equation}

where $\mathrm{NN}_k(x)$ are the nearest k neighbors of $x$. 

(2) \emph{Multi-level Contrastive Learning}

\textbf{Utterance-level:} For each input sentence, we first count all the tokens contained under its slot category, and generate a slot dictionary whose key is all the slot types, and the value corresponding to each key is all the tokens contained under this slot category, which we define as a token starting from B- and ending with the last I- of a phrase or a single word. After the above operation, we can use this dictionary to generate the positive examples of the sentence. We believe that for a given sentence, for each non-O slot, using other tokens under the same slot category to replace the original token in that slot, the multi-intent prediction and slot prediction results for this sentence remain unchanged, therefore, we replace the non-O slot token of this sentence to generate positive examples of the original input sentence, while all other sentences are negative examples of this sentence. In this way, the model can better learn the features of the sentences corresponding to multiple intents, and similar sentence composition can also provide help to predict multiple intents for each other.

\textbf{Slot-level: }For slot-level contrastive learning, we first explain that the unit of comparison at this level is a TOKEN, i.e., a single word or a group of words belonging to a slot label starting with B- and ending with the last I-. So, we define the positive case of TOKEN in each slot label as the phrase that replaces it to generate the positive case of the sentence, that is, the TOKEN randomly selected from the slot dictionary, and the negative case is defined as the TOKEN in other slot labels. This procedure guides the model to better learn the features belonging to the tokens in each slot label, which makes their distribution in feature space more regular. The slot-level contrastive learning loss is computed as follows:
\begin{equation}
\resizebox{.98\hsize}{!}{ $\begin{array}{l}
\mathcal{L}_{\text {SCL}} \\
=-\frac{1}{N_S} \sum_{i=1}^{N_S} \sum_{j=1}^{T_{xS}} \log \frac{e^{\operatorname{sim}\left({x_j}^{(i)} \cdot p o s_{x_j}^{(i)} / \tau\right)}}{e^{\operatorname{sim}\left({x_j}^{(i)} \cdot p o s_{x_j}^{(i)} / \tau\right)}+\sum_{k=1, k \neq i}^{K_S} e^{\operatorname{sim}\left({x_j}^{(i)} \cdot x_{j}^{k} / \tau\right)}}
\end{array}$}
\end{equation}
where $N_S$ denotes the number of slots, $T_{xS}$ denotes the number of positive pairs, $K_S$ denotes the number of negative pairs, and $\tau$ denotes the temperature parameter.

\textbf{Word-level:} Unlike slot-level contrastive learning, the unit of execution for word-level contrastive learning is a word, so we need to think about how to reasonably determine the positive and negative examples of a word. For a single word, it has more valuable information interaction with other words in the same slot. For a word with a non-O slot, its positive example is another word with the same slot suffix. Its negative example is a word with a different slot suffix. For words that cannot be classified into any slots (Eg. OTHERS/UNK), their slot is \textbf{O}, so they will not be replaced or be used to replace other words. Word-level contrastive learning serves as a great guide for generating more relevant slots between similar words, which can be formulated as follows:
\begin{equation}
\resizebox{.98\hsize}{!}{ $\begin{array}{l}
\mathcal{L}_{\text {WCL}} \\
=-\frac{1}{N_W} \sum_{i=1}^{N_W} \sum_{j=1}^{T_{xW}} \log \frac{e^{\operatorname{sim}\left({x_j}^{(i)} \cdot p o s_{x_j}^{(i)} / \tau\right)}}{e^{\operatorname{sim}\left({x_j}^{(i)} \cdot p o s_{x_j}^{(i)} / \tau\right)}+\sum_{k=1, k \neq i}^{K_W} e^{\operatorname{sim}\left({x_j}^{(i)} \cdot x_{j}^{k} / \tau\right)}}
\end{array}$}
\end{equation}
where $N_W$ denotes the number of words, $T_{xW}$ denotes the number of positive pairs, $K_W$ denotes the number of negative pairs, and $\tau$ denotes the temperature parameter.

(3) \emph{Multi-grain Contrastive Learning}  

For utterance-level contrastive learning, we call this coarse-grained contrastive learning if it is performed on a whole-sentence unit, which is calculated as follows:
\begin{equation}
\resizebox{.85\hsize}{!}{$  \begin{array}{l}
\mathcal{L}_{\text {CUCL}} \\
=-\frac{1}{N_C} \sum_{i=1}^{N_C} \log \frac{e^{s i m\left({x}^{i} \cdot {x}^{j} / \tau\right)}}{ e^{si m\left({x}^{i} \cdot {x}^{j} / \tau\right)}+\sum_{k=1}^{K_C} e^{s i m\left({x}^{i} \cdot {x}^{k} / \tau\right)}}
\end{array}$}
\end{equation}
where $N_C$ denotes the number of sentences, $K_C$ denotes the number of negative pairs, and $\tau$ denotes the temperature parameter.

We know that this coarse-grained contrastive learning will undoubtedly help in predicting the intent of the whole sentence for the positive examples we generate, but at the same time, we also expect the utterance-level prediction to contribute to the slot in terms of combining the whole-sentence information as well. Therefore, in order to better learn the frame structure of the sentence, we also perform fine-grained learning at the utterance level. This fine-grained learning is to make the unit of comparison smaller and refine it to each token, and the two levels of contrastive loss formulae are calculated as follows: 
\begin{equation}
\resizebox{.98\hsize}{!}{ $\begin{array}{l}
\mathcal{L}_{\text {FUCL}} \\
=-\frac{1}{N_F} \sum_{i=1}^{N_F} \sum_{j=1}^{T_{xF}} \log \frac{e^{\operatorname{sim}\left({x_j}^{(i)} \cdot p o s_{x_j}^{(i)} / \tau\right)}}{e^{\operatorname{sim}\left({x_j}^{(i)} \cdot p o s_{x_j}^{(i)} / \tau\right)}+\sum_{k=1, k \neq i}^{K_F} e^{\operatorname{sim}\left({x_j}^{(i)} \cdot x_{j}^{k} / \tau\right)}}
\end{array}$}
\end{equation}
where $N_F$ denotes the number of words, $T_{xF}$ denotes the number of positive pairs, $K_F$ denotes the number of negative pairs, and $\tau$ denotes the temperature parameter.

For fine-grained learning, we are more concerned with the frame structure of the whole sentence. Therefore, the relationship between the categories learned in this way is the information that includes both intent information and slot information, and such multi-grained learning can make up for the capture of slot information by utterance-level contrastive learning.
\subsection{Multi-intent Detection Decoder}\label{Multi-intent Detection Decoder}
To better implement multiple intent detection, we conduct a token-level multi-label intent detection, which predicts multiple intents on each token and applies the token-level voting method~\cite{qin2021gl} to select the final output intent. Specifically, we firstly feed the output of self-attentive encoder $\boldsymbol{E}$ = $\{\boldsymbol{e}_{1},\ldots,\boldsymbol{e}_{n}\}$  into an intent-aware BiLSTM:
\begin{equation}
\boldsymbol{h}_{t}^{I}=\operatorname{BiLSTM}\left(\boldsymbol{e}_{t}, \boldsymbol{h}_{t-1}^{I}, \boldsymbol{h}_{t+1}^{I}\right)
\end{equation}
Then we get the intent results $I_{t}$ of the $t$-th word:
\begin{equation}
{I}_t \!=\! \sigma(\boldsymbol{W}_{I}(\operatorname{LeakyReLU}(\boldsymbol{W}_{h}~\boldsymbol{h}_t^{I} \!+\! \boldsymbol{b}_h))\!+\! \boldsymbol{b}_I)
\end{equation}
where $\sigma$ denotes the sigmoid activation function; $\boldsymbol{ W}_{h}$ and $\boldsymbol{ W}_{I}$ are two trainable matrix parameters. If more than half of the tokens predict a value greater than the given threshold for the intent, then the output multiple intents include that intent. In this way, we obtain the predicted utterance-level intent labels $\boldsymbol{o^I}=\{{{o}}_{1}^{I}, {{o}}_{2}^{I}, \ldots, {{o}}_{m}^{I}\}$.

\subsection{Global Graph Interaction Layer}\label{Global Graph Interaction Layer}
Following previous works~\citep{xing2022co,xing2022group}, we utilize a BiLSTM to produce the hidden representation $\boldsymbol{S}$ = ($\boldsymbol{s}_{1}$, \dots, $\boldsymbol{s}_{n}$). At each decoding step $t$, the decoder state ${\boldsymbol{s}}_{t}$ is calculated as follows:
\begin{equation}
{\boldsymbol{s}}_{t} = \operatorname{BiLSTM} \left({{\boldsymbol{I}}_{t} \mathop{||} {\boldsymbol{e}}_{t}, \boldsymbol{s}}_{t-1}, {\boldsymbol{s}}_{t+1} \right)
\end{equation}

where ${\boldsymbol{e}}_{t}$ denotes the aligned encoder hidden state and $\boldsymbol{I}_{t} $ denotes the predicted intent information. Following \cite{qin2021gl}, we continue to use a non-autoregressive slot-filling detection model, but we eliminate the local slot-aware graph interaction layer, which is used for alleviating the uncoordinated slots problem. Comparatively, we use the contrastive learning layer to capture the slot's local dependencies, so we only use the global graph here for the information guidance. We have $n+m$ nodes in the graph where $n$ is the sequence length and $m$ is the number of intent labels predicted by the intent decoder.  The input of slot token feature is ${\boldsymbol{G}^{[S, 1]}}$ = $\{\boldsymbol{s}_{1}, \boldsymbol{s}_{2}, \ldots,\boldsymbol{s}_{n}\} $ and the input intent feature is the embedding ${\boldsymbol{G}^{[I, 1]}}$ =  $\{\phi^{emb}(\boldsymbol{o}_{1}^{I}), \ldots, \phi^{emb}(\boldsymbol{o}_{m}^{I})\}$ where $\phi^{emb}$ is a trainable embedding matrix. There are three types of connections in this graph network, including intent-slot connection, slot-slot connection, and intent-intent connection. The first layer state vector for slot and intent nodes is $\boldsymbol{G}^{1} = \{{\boldsymbol{G}^{[I, 1]}}, {\boldsymbol{G}^{[S, 1]}} \}$. The aggregation process is:
\begin{equation}
\small
{\boldsymbol{g}_{i}^{[S, l+1]}=\sigma\left(\sum_{j \in \mathcal{G}^{S}} \alpha_{i j} \boldsymbol{W}_{g} \boldsymbol{g}_{j}^{[S, l]}+\sum_{j \in \mathcal{G}^{I}} \alpha_{i j} \boldsymbol{W}_{g} \boldsymbol{g}_{j}^{[I, l]}\right)} 
\end{equation}
where $\mathcal{G}^{S}$ and $\mathcal{G}^{I}$ are vertices sets that denote the connected slots and intents respectively. $\boldsymbol{W}_{g}$ is a trainable weight matrix and $\alpha_{i j}$ is the normalized attention coefficients.

After propagation of $L$ layers, we obtain the final slot representation $\boldsymbol{G}^{[S,L+1]}$ for slot prediction:
\begin{equation}
\boldsymbol{y}_{t}^{S}=\operatorname{softmax}\left(\boldsymbol{W}_{s} \boldsymbol{g}_{t}^{[S, L+1]}\right)
\end{equation}
\begin{equation}
\boldsymbol{o}_{t}^{S}=\arg \max \left(\boldsymbol{y}_{t}^{S}\right)
\end{equation}
where $\boldsymbol{W}_{s}$ is a trainable matrix and $\boldsymbol{o}_{t}^{S}$ is the predicted slot of the \textit{t}-th token in the utterance.
\subsection{Self-distillation Layer}\label{Self-distillation Layer}  
In the MMCL model, we designed the self-distillation method to improve the learning ability of the model. Self-distillation minimizes the KL deviation between the current prediction and the previous prediction, specifically, we denote $ p^t_i=P(y^i | x^i, t)$ as the probability distribution of data xi predicted by the model at the $t$-th calendar element with a loss function expressed as:
\begin{equation}
\small
{ \mathcal{L}_{SKL}=\frac{1}{N} \sum_{i}^{N} K L\left(p_{i}^{S, t-1} \mathop{||} p_{i}^{S,t}\right)=\sum_{i}^{N} p_{i}^{S, t-1} \ln \frac{p_{i}^{S, t-1}}{p_{i}^{S,t}}} 
\end{equation}
\begin{equation}
\small
{
  \mathcal{L}_{IKL}=\frac{1}{N} \sum_{i}^{N} K L\left(p_{i}^{I, t-1} \mathop{||} p_{i}^{I,t}\right)=\sum_{i}^{N} p_{i}^{I, t-1} \ln \frac{p_{i}^{I, t-1}}{p_{i}^{I,t}}}
\end{equation}
where $p_i^{S,t}$ denotes the probability distribution of slot, $p_i^{I,t}$ denotes the probability distribution of intent. Note that $p_i^{S,0}$ denotes the one-hot vector of the slot label of $y_i$ and $p_i^{I,0}$ denotes the one-hot vector of the intent label of $y_i$.

Motivated by the success of multi-task learning in various tasks \cite{hu2023scalable,chai2023improving,wang2024discover,hu2024semharq}, we derive the overall training loss of MMCL as below:
\begin{equation}
\begin{array}{cc}
     \mathcal{L}=&\mathcal{L}_{I}+\mathcal{L}_{S}+\mathcal{L}_{CUCL}+\mathcal{L}_{FUCL}   \\
      &+\mathcal{L}_{SCL}+\mathcal{L}_{WCL}+ \mathcal{L}_{SKL}+\mathcal{L}_{IKL}
\end{array}
\end{equation}
\section{Experiments}
\subsection{Datasets and Metrics}
\begin{table*}[h]
	\centering
 	\caption{Results comparison. $\dag$ denotes our model  outperforms baselines with $p<0.01$ under t-test.
	} 
	\begin{adjustbox}{width=.8\textwidth}
		\begin{tabular}{l|ccc|ccc}
			\hline
			\multirow{2}{*}{\textbf{Model}} & \multicolumn{3}{c|}{\textbf{MixATIS}} & \multicolumn{3}{c}{\textbf{MixSNIPS}} \\
			\cline{2-7}
			& Overall(Acc) & Slot(F1) & Intent(Acc) & Overall(Acc) & Slot(F1)& Intent(Acc) 
			\\
			\hline
Attention BiRNN \cite{liu2016attention} &  39.1 &  86.4      &   74.6      & 59.5        &  89.4     & 95.4 \\
Slot-Gated \cite{goo2018slot}    &  35.5 &  87.7      &   63.9      & 55.4        &  87.9     & 94.6 \\
Bi-Model \cite{wang2018bi}         &  34.4 &  83.9      &   70.3      & 63.4        &  90.7    & 95.6 \\
SF-ID \cite{haihong2019novel}               &  34.9 &  87.4      &   66.2      & 59.9        &  90.6     & 95.0 \\
Stack-Propagation \cite{qin2019stack}&  40.1 &  87.8      &   72.1      & 72.9        &  94.2     & 96.0 \\
Joint Multiple ID-SF \cite{gangadharaiah2019joint}&36.1 &84.6    &   73.4      & 62.9        &  90.6     & 95.1 \\
AGIF \cite{qin2020agif}                &  40.8 &  86.7      &   74.4      & 74.2        &  94.2     & 95.1 \\
LR-Transformer \cite{cheng2021result, cheng2021effective}    &  43.3 &  88.0      &   76.1      & 74.9        &  94.4     & 96.6 \\ 
GL-GIN \cite{qin2021gl}        &  43.0 &  88.2      &   76.3      & 73.7        &  94.0     & 95.7 \\ 
SDJN \cite{chen2022joint}    &    44.6    &  88.2      &   77.1    & 75.7        &  94.4     & 96.5 \\ 
GISCo \cite{Enhancing}  &  48.2 &  88.5      &   75.0      & 75.9        &  95.0     & 95.5 \\ 
Co-guiding Net\cite{xing2022co}    &  51.3 &  89.8      &   79.1      & 77.5        &  95.1     & 97.7 \\ 
ReLa-Net \cite{xing2022group}    &  52.2 &  90.1      &   78.5     & 76.1        &  94.7     & 97.6 \\ 
SSRAN      \cite{cheng2022scope}  &  48.9 &  89.4     &   77.9      & 77.5        &  95.8     & 98.4 \\ 
\hline
MMCL (ours)           &  \textbf{54.8}$^\dag$ &\textbf{92.5}$^\dag$ &\textbf{81.4}$^\dag$ & \textbf{79.5}$^\dag$  &  \textbf{96.3}$^\dag$ & \textbf{98.6}$^\dag$ \\
			\hline 
		\end{tabular}
	\end{adjustbox}
	\label{tab:main results}
\end{table*}
\begin{table*}[t] 
	\small
	\centering
 	\caption{Results of ablation experiments on the MixATIS and the MixSNIPS datasets.
	} 
	\begin{adjustbox}{width=\textwidth}
		\begin{tabular}{l|ccc|ccc}
			\hline
			\multirow{2}{*}{\textbf{Model}} & \multicolumn{3}{c|}{\textbf{MixATIS}} & \multicolumn{3}{c}{\textbf{MixSNIPS}} \\
			\cline{2-7}
			& Overall(Acc)& Slot(F1) & Intent(Acc) & Overall(Acc) & Slot(F1)  & Intent(Acc) 
			\\
			\hline
MMCL          &  \textbf{54.8} &\textbf{92.5} &\textbf{81.4} & \textbf{79.5}  &  \textbf{96.3} & \textbf{98.6} \\\hline
w/o  Word-level CL  &  53.0 ($\downarrow$1.8) &  89.8 ($\downarrow$2.7)      &   79.5    ($\downarrow$1.9)  & 77.4 ($\downarrow$2.1)       &  95.1 ($\downarrow$1.2)    & 98.0 ($\downarrow$0.6) \\
w/o  Slot-level CL &  52.6 ($\downarrow$2.2)&  89.1  ($\downarrow$3.4)    &   79.3  ($\downarrow$2.1)    & 76.8    ($\downarrow$2.7)    &  94.9  ($\downarrow$1.4)  & 97.7 ($\downarrow$0.9)\\\hdashline
w/o Utterance-level CL &  52.0 ($\downarrow$2.8) &  90.4  ($\downarrow$2.1)  & 78.5 ($\downarrow$2.9)   & 76.4       ($\downarrow$3.1) &  95.2    ($\downarrow$1.1) & 97.4 ($\downarrow$1.2) \\
w/o Utterance-level Coarse-grain CL &  52.6 ($\downarrow$2.2) &  91.3  ($\downarrow$1.2)  & 78.8 ($\downarrow$2.6)   & 76.9       ($\downarrow$2.6) &  95.9    ($\downarrow$0.4) & 97.8 ($\downarrow$0.8) \\
w/o Utterance-level Fine-grain CL &  52.8 ($\downarrow$2.0) &  90.8  ($\downarrow$1.7)  & 80.6 ($\downarrow$0.8)   & 77.2      ($\downarrow$2.3) &  95.5    ($\downarrow$0.8) & 98.2 ($\downarrow$0.4) \\\hdashline
w/o Self-distillation & 53.2  ($\downarrow$1.6) &  91.6  ($\downarrow$0.9)  &   80.4 ($\downarrow$1.0) & 78.2   ($\downarrow$1.3)     &  95.6 ($\downarrow$0.7)    & 97.8 ($\downarrow$0.8)  \\
w/o Intent Self-distillation & 53.6  ($\downarrow$1.2) &  92.2  ($\downarrow$0.3)  &   80.8 ($\downarrow$0.6) & 78.6   ($\downarrow$0.9)     &  96.1 ($\downarrow$0.2)    & 98.0 ($\downarrow$0.6)  \\
w/o Slot Self-distillation & 53.9  ($\downarrow$0.9) &  91.8  ($\downarrow$0.7)  &   81.1 ($\downarrow$0.3) & 78.8   ($\downarrow$0.7)     &  95.8 ($\downarrow$0.5)    & 98.2 ($\downarrow$0.4)  \\\hdashline
+ Local Slot-aware GAT & 54.5  ($\downarrow$0.3) &  92.1  ($\downarrow$0.4)  &   81.2 ($\downarrow$0.2) & 77.7   ($\downarrow$1.8)     &  95.7 ($\downarrow$0.6)    & 97.0 ($\downarrow$1.6)  \\
			\hline
		\end{tabular}
	\end{adjustbox}
	\label{tab:ablation}
\end{table*}
Two public datasets\footnote{\url{https://github.com/LooperXX/AGIF}} are used for evaluating our model, which is the cleaned version of MixATIS and MixSNIPS~\cite{qin2020agif}. MixATIS dataset is collected from ATIS dataset~\cite{hemphill1990atis} and MixSNIPS dataset is collected from SNIPS dataset~\cite{coucke2018snips}. Compared to the single-domain MixATIS dataset, the MixSNIPS dataset is more complicated mainly due to the intent diversity and larger vocabulary.

For all the experiments, we select the trained model that works the best on the \textit{dev} set and then evaluate it on the \textit{test} set. To avoid overfitting, the training will early-stop if the loss on \texttt{dev} set does not decrease for three epochs. We evaluate the performance of slot filling with F1 score, intent detection with accuracy, and the NLU semantic frame parsing with overall accuracy that represents both the metrics that are correct in the utterance.

\subsection{Experimental Settings}
The dimensionality of the embedding is 128 and 64 on Mix-ATIS and Mix-SNIPS, respectively. The dimensionality of the LSTM hidden units is 256. The batch size is 16. The number of the multi-head is 4 and 8 on the MixATIS and MixSNIPS datasets, respectively. The layer number of the graph attention network is set to 2. The temperature parameter $\tau$ is set to 2. The value of label smoothing is set to 0.1. Sign test~\cite{collins2005clause} is a standard statistical-significance test. Following previous works~\cite{qin2020agif,qin2021gl,cheng2023mrrl,zhu2023enhancing}, the word and label embeddings are trained from scratch. We conduct all experiments on a single Nvidia V100 GPU.

\subsection{Main Results}
Experimental results are shown in Table \ref{tab:main results}. Based on them, we have the following observations: 
\begin{figure*}[t]
  \centering
  \includegraphics[width=.9\linewidth]{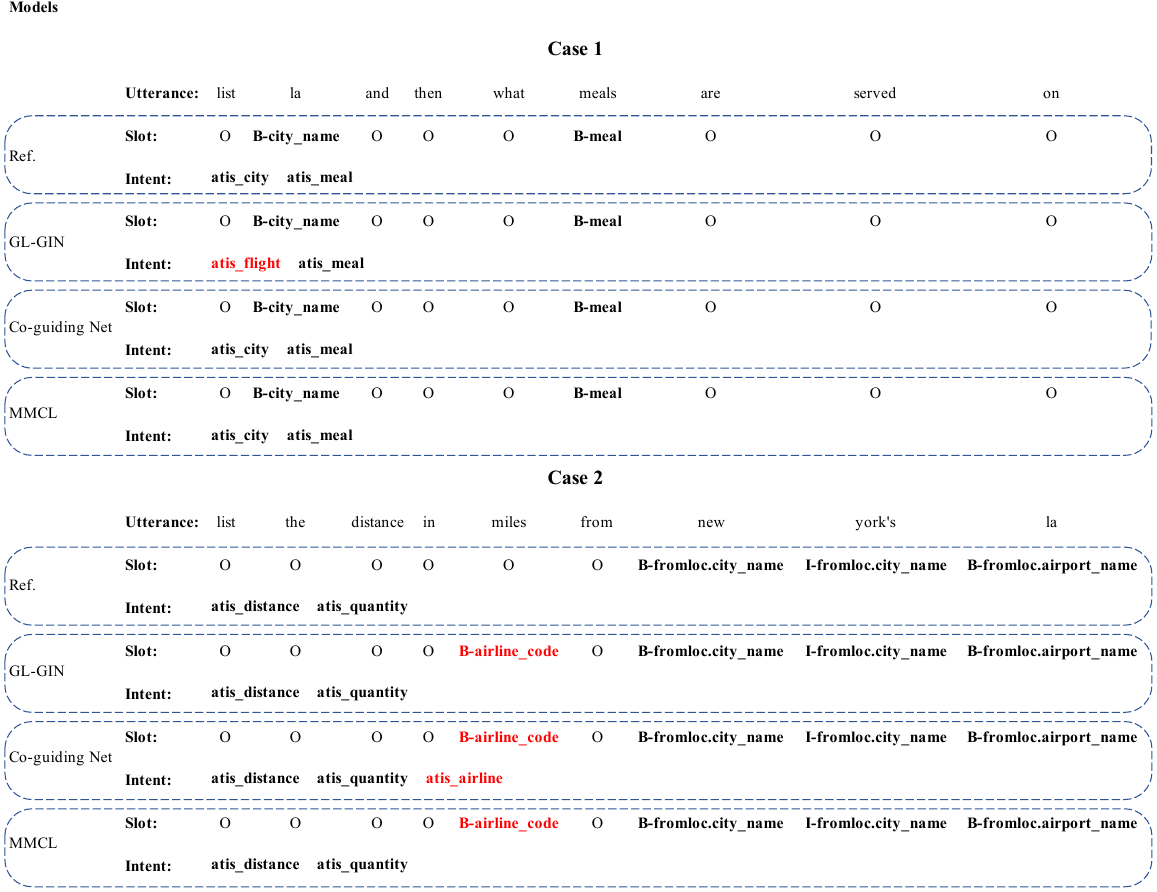}
  \caption{Case study of our model compared to previous models in achieving mutual guidance and avoiding error propagation.}
  \label{fig:casestudy}
\end{figure*}

(1)~MMCL achieves significant and consistent improvements across all tasks and datasets. This is because our model achieves mutual guidance between the multi-intent detection and slot-filling tasks through multi-level comparison and multi-granularity comparison, allowing both tasks to provide critical cues for each other.

(2)~Our model obtained higher results on the prediction of multiple intentions. Our model introduces a multi-level contrastive learning layer, compared to the previous model, through which the model learns the relationship between the individuals of that level in the representation space, allowing individuals with mutually supportable properties to help each other. In overall multi-level structure, the model has a deeper exploration of the structure of the sentence, thus deepening the relationship between the two tasks of intent detection and slot filling. In addition, the multi-grain exploration at the utterance level also plays a significant role in the improvement of multi-intent results.

(3)~Our model improves the overall results. Due to the coupling of intent detection and slot filling, the more we can use their relationship to guide each other's predictions, the more likely we are to get better overall results. Multi-level contrastive learning does this well, i.e., it does not lead to error propagation as multi-stage cross-training may do, and it takes better care of the semantic structure at each level relative to other joint models.
\subsection{Model Analysis}
To verify the advantages of MMCL from different perspectives, we conduct ablation experiments on MixATIS and MixSNIPS.

(1) \emph{Effect of Mulit-level Contrastive learning}.

\textbf{Utterance-level:} To demonstrate the effectiveness of utterance-level contrastive learning, we designed a variant termed w/o Utterance-level CL, and the results are shown in Table \ref{tab:ablation}. We can see that after we remove the utterance-level contrastive learning, the Intent Acc of MixATIS decreases by 2.9$\%$ and MixSNIPS decreases by 1.2$\%$. In addition, the overall Acc drop is even more significant: 2.8$\%$ for MixATIS and 3.1$\%$ for MixSNIPS. This demonstrates that learning from utterance-level contrast can effectively contribute to achieving mutual guidance between the two tasks and can improve the overall Acc. This level of comparison can be a preliminary interaction learning for multiple intents as well as slot filling. For sentences with similar semantic frame structures, their corresponding intent prediction and slot prediction for each word can be cross-guided.
\begin{figure*}[t]
  \centering
  \includegraphics[width=.8\linewidth]{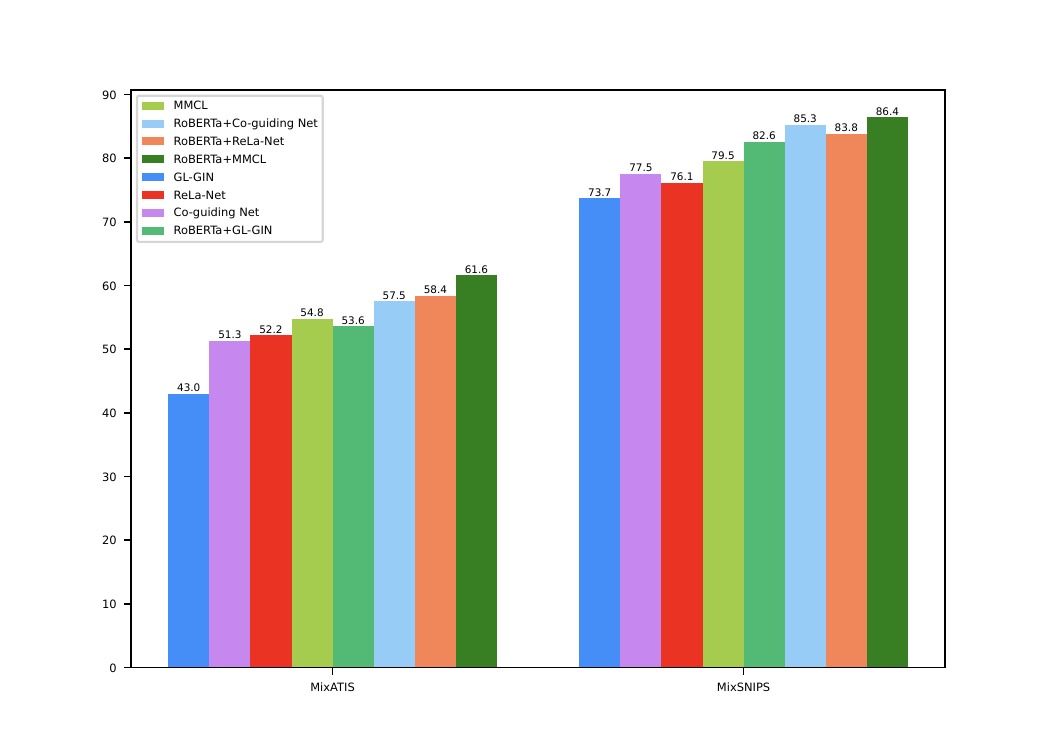}
  \caption{Overall accuracy comparison with RoBERTa.}
  \label{fig:PLM}
\end{figure*}

\textbf{Slot-level:} We also explore the effectiveness of slot-level contrastive learning on the model, we designed a variant termed w/o  Slot-level CL, whose results are shown in Table \ref{tab:ablation}. We can see that after we remove the contrastive learning at the slot level, MixATIS has a 3.4$\%$ drop in slots and MixSNIPS has a 1.4$\%$ drop. In addition, the overall Acc drop is even more significant: 2.2$\%$ for MixATIS and 2.7$\%$ for MixSNIPS. This demonstrates that this level of contrastive learning learns slot prediction in a specific semantic context by forming positive examples for slots between positive sentences.

\textbf{Word-level:} Finally, we want to test the role of word-level contrastive learning in our model, which we believe further enhances the results of the experiment based on the first two levels. We remove this level by a variant termed w/o  Word-level CL, and the result is shown in Table \ref{tab:ablation}, we can see that in combination with the upper two layers, it maximizes the intrinsic associations learned by contrastive learning. After removing it, MixATIS has a 2.7$\%$ drop in slots and MixSNIPS has a 1.2$\%$ drop. We use word-level contrastive learning to strengthen the connection between words that belong to the same slot of the same label, and thus strengthen the guidance of slot-filling prediction and intention prediction for such words. This layer of contrastive learning guides the model to learn more detailed slot boundaries based on the previous two layers, which further illustrates the need for a multi-level contrastive structure.

(2) \emph{Effect of Multi-grain Contrastive learning}.
 We designed a variant termed Utterance-level Coarse-grain CL and a variant termed Utterance-level Fine-grain CL to verify the validity of multi-grain contrastive learning. Without Coarse-grain CL, overall, intent acc, and F1 slot dropped by 2.2$\%$,1.2$\%$,2.6$\%$ respectively on dataset MixATIS, and 2.6$\%$,0.4$\%$,0.8$\%$ on dataset MixSNIPS. We can see that the results with coarse granularity removed are worse than those with fine granularity removed in terms of intent, due to the fact that coarse granularity focuses more on the completeness of a whole sentence compared to fine granularity, which facilitates the identification of intent.
 Correspondingly, the results with coarse granularity removed are better than those with fine granularity removed in terms of slotting, due to the fact that fine granularity can disassemble the sentence frame and learn the slotting information.

(3) \emph{Effect of Self-distillation}.
 We designed the removal of the entire self-distillation module, the intent detection distillation module, and the slot filling distillation module, respectively,  we can find there was a different degree of decline in the results, so the self-distillation in our task has played a certain effect. Through our observation of the training data, we found that some data have controversial labels. We call it label noise, i.e., different people have different decisions on the slot of the same word, and so does the intent of the same sentence, which may be noisy, and we solve this problem by distillation to a certain extent. Notably, the results of our model with the entire distillation part removed are still higher than those of the baseline, which further shows the superiority of our proposed multi-level and multi-grain contrastive learning framework.

(4) \emph{Case Study}.
As shown in Figure \ref{fig:casestudy}, we provide case studies that include a comparison of two sentence sequences and their prediction results from different models. In the first case, we can observe that GL-GIN incorrectly predicts the intents, but both Co-guiding Net and MMCL correctly predict the intents. One possible reason is that both Co-guiding Net and MMCL achieve mutual guidance between intent and slot, which can further improve performance. In the second case, we can observe that all the models incorrectly predict the slots but only Co-guiding Net incorrectly predicts the intents. One possible reason is that Co-guiding Net suffers from error propagation. In contrast, MMCL avoids the problem, which shows the superiority of MMCL.

(5) \emph{Results on Pre-trained Language Model}.
Pre-trained language models have already demonstrated exceptional performance across numerous tasks~\cite{xin2023pooling,li2023pace,lin2023consistent123,li2023vision,wang2023out,guo2023focus,li2023learning,xin23c_interspeech,huang2023iteratively,xin23d_interspeech}. To explore the effect of the pre-trained language model, we use a pre-trained RoBERTa~\cite{liu2019roberta} encoder to replace the self-attentive encoder and RoBERTa is fine-tuned in the training process. For each word, the hidden state of its first subword at the last layer is taken as its word representation fed to the BiLSTMs in Sec. \ref{Self-Attentive Encoder}. Figure \ref{fig:PLM} shows the results comparison of our MMCL, GL-GIN, Co-guiding Net, and ReLa-Net as well as their variants using RoBERTa encoder. We can find that the pre-trained RoBERTa encoder can bring remarkable improvements by generating word representations with high quality. Moreover, RoBERTa+MMCL reaches a new state-of-the-art performance on the two datasets, which also verifies the effectiveness of our method.
\vspace{-2mm}
\section{Conclusion}
In this paper, we propose MMCL, a novel framework for the multi-intent SLU task, which applies contrastive learning at three levels to enable intent and slot to mutually guide each other. In addition, we apply a self-distillation method to improve the robustness of the model by minimizing the KL deviation between the current prediction and the previous prediction. Experiments and further analysis on two benchmark datasets show that our model outperforms previous models and achieves new state-of-the-art performance. In the future, we will continue to explore how to achieve mutual guidance between slot and intent for multi-intent SLU.
\begin{acks}
We thank all the anonymous reviewers for their insightful comments. This paper was partially supported by Shenzhen Science \& Technology Research Program (No:GXWD20201231165807007-20200814115301001) and NSFC (No: 62176008).
\end{acks}
\bibliographystyle{unsrt}
\balance
\bibliography{sample-base}
\end{document}